# Implementing graph neural networks with TensorFlow-Keras

*Patrick Reiser[1,2], Andre Eberhard[2] and Pascal Friederich[1,2]*

[1]Institute of Nanotechnology, Karlsruhe Institute of Technology (KIT), Hermann-von-Helmholtz-Platz 1, 76344 Eggenstein-Leopoldshafen,Germany
[2]Institute of Theoretical Informatics, Karlsruhe Institute of Technology (KIT), Am Fasanengarten 5, 76131 Karlsruhe, Germany

E-Mail: patrick.reiser@kit.edu, pascal.friederich@kit.edu

Graph neural networks are a versatile machine learning architecture that received a lot of attention recently. In this technical report, we present an implementation of convolution and pooling layers for TensorFlow-Keras models, which allows a seamless and flexible integration into standard Keras layers to set up graph models in a functional way. This implies the usage of mini-batches as the first tensor dimension, which can be realized via the new RaggedTensor class of TensorFlow best suited for graphs. We developed the **K**eras **G**raph **C**onvolutional **N**eural **N**etwork Python package ***kgcnn*** based on TensorFlow-Keras that provides a set of Keras layers for graph networks which focus on a transparent tensor structure passed between layers and an ease-of-use mindset.

## Introduction

Graph neural networks (GNNs) are a natural extension of common neural network architectures like convolutional neural networks (CNN) for image classification to graph structured data.[1] For example, recurrent,[2,3] convolutional,[1,4–6] and spatial-temporal[7] graph neural networks as well as graph autoencoders[8,9] and graph transformer models[10,11] have been reported in literature. A graph $G = (V, E)$ is defined as a set of vertices or nodes $v_i \in V$ and edges $e_{ij} = (v_i, v_j) \in E$ connecting two nodes. There are already comprehensive and extensive review articles for graph neural networks, which summarize and categorize relevant literature on graph learning.[12] The most frequent applications of GNNs are either node classification or graph embedding tasks. While node classification is a common task for very large graphs such as citation networks[9] or social graphs,[13] graph embedding learns a representation of smaller graphs like molecules[4] or text classifications.[14,15] Graph convolutional neural networks (GCN) stack multiple convolutional and pooling layers for deep learning to generate a high-level node representation from which both a local node and global graph classification can be obtained. Most GCNs can be considered as message passing networks,[16] where neighbouring nodes propagate information between each other along edges. In each update step $t$, the central nodes hidden representation $h_v$ is convolved with its neighbourhood given by:

$$h_v^{t+1} = U_t(h_v^{t+1}, m_v^{t+1}),$$

where $m_v^t$ denotes the aggregated message and $U_t$ the update function. The message to update is usually (more complex aggregation is of course possible) acquired from summing message functions $M_t$ from the neighbourhood $N(v) = \{u \in V \mid (u, v) \in E\}$ of node $v$:

$$m_v^{t+1} = \sum_{w \in N(v)} M_t(h_v^t, h_w^t, e_{vw}).$$

There is a large variety in convolution operators, which can be spectral-based[17] or spatial-based involving direct neighbours or a path of connected nodes to walk and collect information from.[11,18] Moreover, the message and update functions can be built from recurrent networks,[19] multi-layer perceptrons (MLP)[20] or attention heads[21] which are complemented by a set of possible aggregation or pooling operations. Aggregation is usually done by a simple average of node representations or by a more refined set2set encoder part[22] as proposed by Gilmer et al.[16]. A reduction of nodes in the graph is achieved by pooling similar to CNNs but which is much more challenging on arbitrarily structured data. Examples of possibly differentiable and learnable pooling filters introduced in literature are DiffPool,[23] EdgePool,[24] gPool,[25] HGP-SL,[23] SAGPool,[26] iPool,[27] EigenPool[28] and graph based clustering methods such as the Graclus algorithm.[29–32]

In order to utilize the full scope of different graph operations for setting up a custom GNN model, a modular framework of convolution and pooling layers is necessary. We briefly summarize and discuss existing graph libraries and their code coverage. Then, a short overview of representing graphs in tensor form is given. Finally, we introduce our graph package *kgcnn* for Tensorflow's 2.0 Keras API,[33–35] which seamlessly integrates graph layers into the Keras[36] environment.

## Graph Libraries

Since graph neural networks require modified convolution and pooling operators, many Python packages for deep learning have emerged for either TensorFlow[33,34] or PyTorch[37] to work with graphs. We try to summarize the most notable ones without any claim that this list is complete.

*PyTorch Geometric*.[38] A PyTorch based graph library which is probably the largest and most used graph learning Python package up to date. It implements a huge variety of different graph models and uses a disjoint graph representation to deal with batched graphs (graph representations are discussed in the next section).

*Deep Graph Library (DGL)*.[39] A graph model library with a flexible backend and a performance optimized implementation. It has its own graph data class with many loading options. Moreover, variants such as generative graph models,[40] Capsule[41] and transformers[42] are included.

*Spektral*.[43] A Keras[36] implementation of graph convolutional networks. Originally restricted to spectral graph filters,[17] it now includes spatial convolution and pooling operations. The graph representation is made flexible by different graph modes detected by each layer.

*StellarGraph*.[44] A Keras[36] implementation that implements a set of convolution layers and a few pooling layers plus a custom graph data format.

With PyTorch Geometric and DGL there are already large graph libraries with a lot of contributors from both academics and industry. The focus of the graph package presented here is on a neat integration of graphs into the TensorFlow-Keras framework in the most straightforward way. Thereby, we hope to provide Keras graph layers which can be quickly rearranged, changed and extended to build custom graph models with little effort. This implementation is focused on the new TensorFlow's RaggedTensor class which is most suited for flexible data structures such as graphs and natural language.

## Graph representation

A main issue with handling graphs is their flexible size, which is why graph data can not be easily arranged in tensors as it is done for example in image processing. Especially arranging smaller graphs of different size in mini-batches poses a problem with fixed sized tensors. A way to circumvent this problem is to use zero-padding with masking or composite tensors such as ragged or sparse tensors. Another possibility is to join small graphs into a single large graph, where the individual subgraphs are not connected to each other, which is illustrated in **Figure 1** and is often referred to as disjoint representation. The tensors used to describe a graph are typically given by a node list *n* of shape *([batch], N, F)*, a connection table of edge indices of incoming and outgoing node *m* with shape *([batch], M, 2)* and a corresponding edge feature list *e* of shape *([batch], M, F)*. Here, *N* the number of nodes, *F* denotes the dimension of the node representation and *M* the number of edges. A common representation of a graph's structure is given by the adjacency matrix *A* of shape *([batch], N, N)* which has $A_{ij} = 1$ if the graph has an edge between nodes *i* and *j* and $A_{ij} = 0$ otherwise.

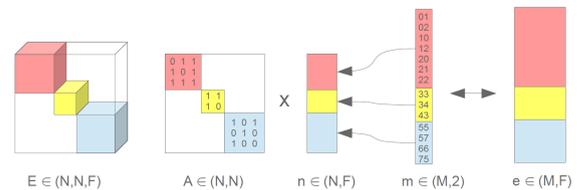

*Figure 1: Disjoint graph representation with adjacency matrix A, node list n and connection table m. Edge features are added in form of a feature list e or a feature matrix E matching A. The indices in m match the total graph as indicated by arrows. The subgraph distinction encoded by color has to be stored separately.*

However, with RaggedTensors, node features and edge index lists can be passed to Keras models with a flexible tensor dimension that incorporates different numbers of nodes and edges. For example, a ragged node tensor of shape *(batch, None, F)* can accommodate a flexible graph size in the second dimension. It is to note that even sparse matrices, which are commonly used to represent the adjacency matrix in a disjoint representation, are internally stored as a value plus index tensor. This means that the ragged tensor representation can be cast into a sparse or padded representation with little cost, if necessary. TensorFlow 2.0 further supports limited sparse matrix operations, which can be used for graph convolution models like GCN.

## Keras graph package - kgcnn

A flexible and simple integration of graph operations into the TensorFlow-Keras framework can be achieved via ragged tensors. As mentioned above, ragged tensors are capable of efficiently representing graphs and have inherently access to various methods within TensorFlow. For more sophisticated pooling algorithms which can not be operated on batches, a parallelization of individual graphs within the batch could be achieved with the TensorFlow map functionality, although this is less efficient than vectorized operations and depends on implementation details.

| Model | HOMO [eV] | LUMO [eV] | $E_G$ [eV] |
|---|---|---|---|
| MPN | 0.061 | 0.047 | 0.083 |
| Schnet | 0.044 | 0.038 | 0.067 |
| MegNet | 0.045 | 0.037 | 0.066 |

*Table 1:* Mean absolute validation error for single training on QM9 dataset for targets like HOMO, LUMO level and gap $E_G$ in eV using popular GNN architectures implemented in kgcnn. No hyperparameter optimization or feature engineering was performed.

Consequently, we introduce a Python package *kgcnn* (https://github.com/aimat-lab/gcnn_keras) that uses RaggedTensors, which are passed between graph layers for graph convolution and message passing models. We believe that the use of RaggedTensors makes it easy to debug code, allows a transparent and readable coding style, and enables a seamless integration with many TensorFlow methods which are available for custom layers. We implemented a set of basic Keras layers for TensorFlow 2.0 from which many models reported in literature can be constructed. The Python package implements as an example: GCN,[1] Interaction network,[6] message passing,[16] Schnet,[4] MegNet[20] and Unet[25]. The focus is set on graph embedding tasks, but also node and link classification tasks can be implemented using *kgcnn*. The models were tested with common bench-mark datasets such as Cora,[45] MUTAG[46] and QM9.[47] Typical benchmark accuracies such as chemical accuracy on the QM9 dataset are achieved with the corresponding models implemented in *kgcnn*.

## Conclusion

In summary, we discussed a way to integrate graph convolution models into the TensorFlow-Keras deep learning framework. Main focus of our *kgcnn* package is the transparency of the tensor representation and the seamless integration with other Keras models. We plan to continue to extend the *kgcnn* library to incorporate new models, in particular GNNExplainer[48] and DiffPool,[23] and improve functionality.

## Code availability

The package is available on the github repository https://github.com/aimat-lab/gcnn_keras and through the Python Package Index via *pip install kgcnn.*

## Acknowledgement

P.F. acknowledges funding from the European Union's Horizon 2020 research and innovation programme under the Marie Sklodowska-Curie grant agreement No 795206.